\documentclass[letterpaper, 10 pt, conference]{ieeeconf}  % Comment this line out if you need a4paper

\IEEEoverridecommandlockouts                              % This command is only needed if 
\title{\LARGE \bf
Ergodic Trajectory Optimization on Generalized Domains Using Maximum Mean Discrepancy 
}

\author{Christian Hughes$^{1}$, Houston Warren$^{2}$, Darrick Lee$^{3}$, Fabio Ramos$^{4}$, and Ian Abraham$^{1}$% <-this % stops a space
\thanks{$^{1}$Christian Hughes and Ian Abraham are with Department of Mechanical Engineering,
        Yale University, New Haven, CT 06520, USA
        {\tt\small \{christian.hughes, ian.abraham\}@yale.edu}}%
\thanks{$^{2}$Houston Warren is with the School of Computer Science,            The University of Sydney,
        Sydney, Australia
        {\tt\small houston.warren@sydney.edu.au}}%
\thanks{$^{3}$Darrick Lee is with the School of Mathematics,
    The University of Edinburgh, 
        Edinburgh, Scotland
        {\tt\small darrick.lee@ed.ac.uk}}
\thanks{$^{4}$Fabio Ramos is with the School of Computer Science, The           University of Sydney,
        Sydney, Australia, and NVIDIA, USA
        {\tt\small fabio.ramos@sydney.edu.au}}%
}

\usepackage{tabularx,booktabs} % For Tables ---------
\usepackage{amsmath, amssymb} % For Math --------
\usepackage{xcolor}
\usepackage{graphicx}
\usepackage{multirow}

\makeatletter
\newcommand{\linebreakand}{%
  \end{@IEEEauthorhalign}
  \hfill\mbox{}\par
  \mbox{}\hfill\begin{@IEEEauthorhalign}
}
\makeatother

\newcommand{\E}{\mathbb{E}}
\newcommand{\bx}{\mathbf{x}}

\newcommand{\cE}{\mathcal{E}}
\newcommand{\cH}{\mathcal{H}}
\newcommand{\MMD}{\text{MMD}}

\newtheorem{theorem}{Theorem}
\newtheorem{definition}{Definition}

\begin{document}

\maketitle
\thispagestyle{empty}
\pagestyle{empty}

\begin{abstract}
    We present a novel formulation of ergodic trajectory optimization that can be specified over general domains using kernel maximum mean discrepancy. 
    Ergodic trajectory optimization is an effective approach that generates coverage paths for problems related to robotic inspection, information gathering problems, and search and rescue. 
    These optimization schemes compel the robot to spend time in a region proportional to the expected utility of visiting that region. 
    % These methods achieve effective coverage by optimizing where a robot spends time, on average, along its trajectory such that the robot spends time in areas proportional to the expected utility of visiting said area. 
    Current methods for ergodic trajectory optimization rely on domain-specific knowledge, e.g., a defined utility map, and well-defined spatial basis functions to produce ergodic trajectories. 
    Here, we present a generalization of ergodic trajectory optimization based on maximum mean discrepancy that requires only samples from the search domain.
    % and yields a metric to generate ergodic trajectories. 
    % Maximum mean discrepancy is a two-sample statistical test that induces a metric on achieving ergodicity that can be used to 
    % compare a robot’s time-average trajectory statistics to arbitrary samples of the utility of areas defined on arbitrary search domains. 
    We demonstrate the ability of our approach to produce coverage trajectories on a variety of problem domains including robotic inspection of objects with differential kinematics constraints and on Lie groups without having access to domain specific knowledge. 
    Furthermore, we show favorable computational scaling compared to existing state-of-the-art methods for ergodic trajectory optimization with a trade-off between domain specific knowledge and computational scaling, thus extending the versatility of ergodic coverage on a wider application domain.
\end{abstract}

\section{Introduction}

    Ergodic trajectory optimization has recently been presented as an effective method for coverage and exploration in problems relating to robotic inspection~\cite{srinivasan2023multi}, information gathering~\cite{dressel2018optimality}, search and rescue~\cite{prabhakar2020ergodic, sartoretti2021spectral}, and tactile exploration~\cite{kalinowska_ergodic_2021, Abraham-RSS-18}.
    Existing approaches produce ergodic coverage trajectories by optimizing the time-averaged trajectory statistics of a robot, i.e., where the robot spends time in a domain, to be proportional to the utility of visiting said area~\cite{mathew2011, miller2013trajectory, lee2024}. 
    Ergodic trajectories have an asymptotic guarantee to completely explore a bounded domain, proportional to the expected utility, as time approaches infinity~\cite{mavrommati2018}. 
    However, existing methods require that the search domain and underlying utility are defined and known~\cite{sun2024fast, mathew2011}, which may not be readily obtainable in practice. 

    In many robotic scenarios, it is often the case that information about the search domain is observed through sensor information, e.g., via depth sensors, and numerically approximated.
    Similarly, in real-time settings, it is necessary to adapt exploration according to spatio-temporal domains that need not have closed-form solutions and require numerical approximations~\cite{lei2022, oh2018, majeed2021}.     
    Methods for ergodic coverage that use numerical approximations of domains via meshes or point clouds show promise in these problem settings~\cite{ivic2022, bilaloglu2024tactile, ivic2023multi}, but are computationally expensive as a result of approximating basis functions and solving a partial differential equation that promotes ergodicity.

\begin{figure}[t!]
    \centering
    \includegraphics[trim={0 10cm 0 0},clip,width=\linewidth]{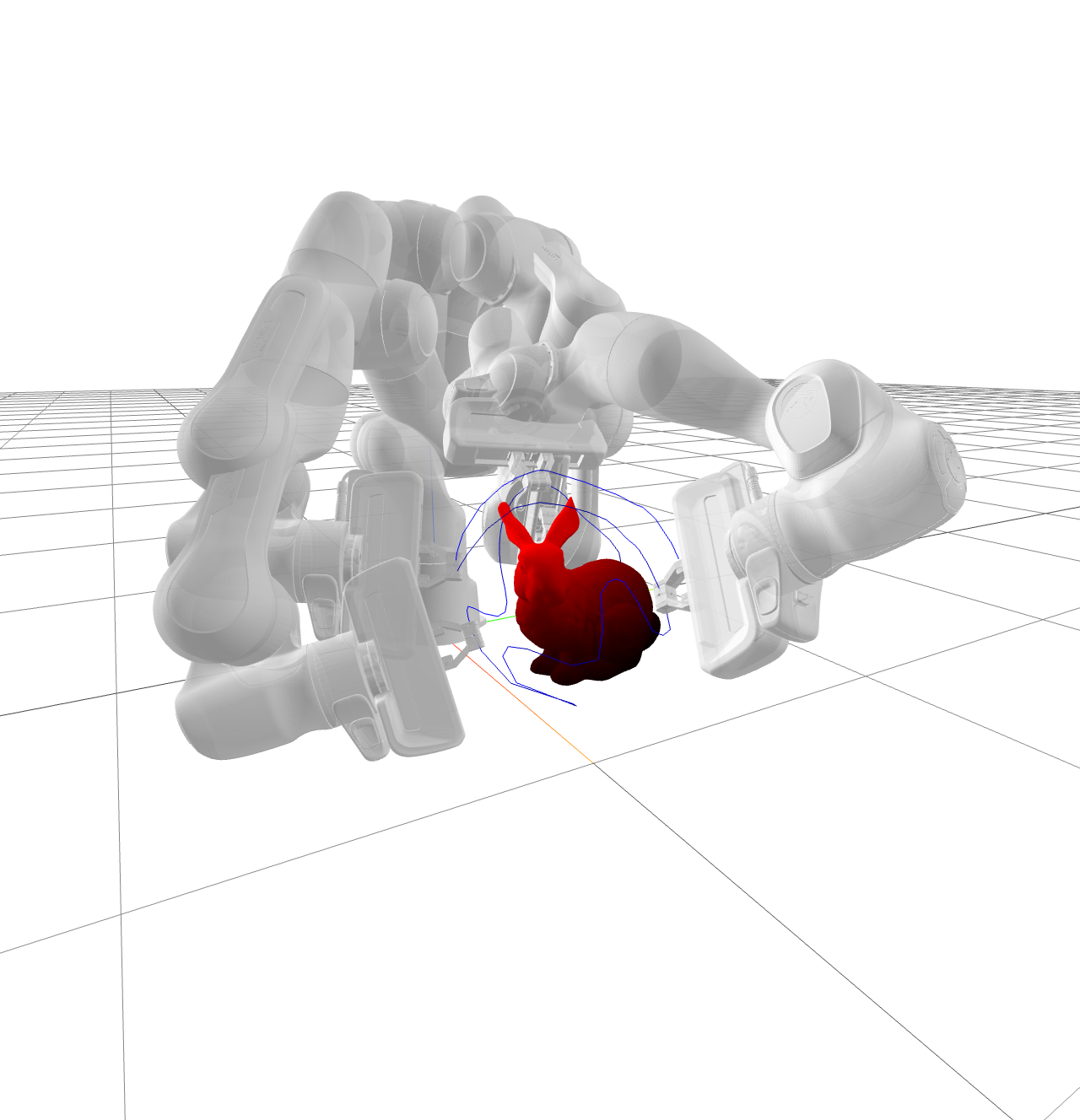}
    \caption{\textbf{Ergodic trajectory optimization via maximum mean discrepancy enables robotic search over arbitrary objects.} Illustration of our approach for generating ergodic trajectories for inspection of the bunny (red areas are of more importance) while respecting differential kinematic constraints of the Franka panda robot. Note only samples from the bunny surface are needed to compute trajectories.}
    \label{fig:enter-label}
\end{figure}

    \begin{figure*}[ht!]
        \centering
        \includegraphics[trim={0 0cm 0 0.5cm},clip,width=\linewidth]{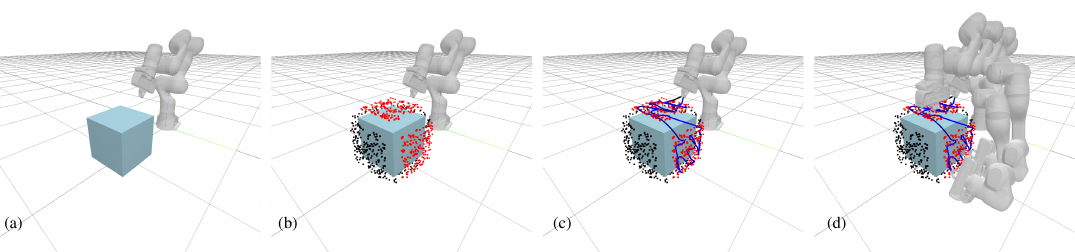}
        \caption{\textbf{Ergodic trajectory optimization via MMD procedure.} (a) A typical robotic inspection setting where the goal is to scan a object (box) while respecting robot constraints. (b) Generated samples (e.g., from a depth camera) with inspection areas of higher importance shown in red. (c) Ergodic trajectories optimized using the ergodic MMD metric~\eqref{eq:mmd_empirical} from samples with differential kinematics constraints (in joint space). Samples are shown offset the box used to implicitly induce collision avoidance constraints. (d) Executed ergodic trajectories inspecting the box uniformly across the two-faces. }
        \label{fig:pipeline}
    \end{figure*}

    In this paper, we circumvent the need for domain specific knowledge through a novel generalization of ergodic trajectory optimization that only requires samples from search domains. 
    Our approach is derived from the kernel maximum mean discrepancy metric (MMD)\footnote{Throughout this article, MMD will always refer to kernel MMD.}, which is a two-sample statistical test that operates only using samples of the relevant distributions to measure similarity~\cite{muandet2017, gretton2012}. 
    We formulate an ergodic metric as the maximum mean discrepancy between the time-averaged trajectory statistics and samples of a utility measure over a search domain. 
    We find our approach is able to optimize for ergodic trajectories across a variety of domains without the need of domain specific knowledge or extensive computation of basis functions and partial differential equations.
    % In fact, we demonstrate that the proposed approach generalizes existing ergodic coverage methods allowing us to be expressive in our coverage and exploration task.
    We showcase our approach on a number of robotic exploration tasks and compare our method to existing ergodic coverage methods while not requiring domain specific information.
    % inspection, and information gathering problems. 
    % Furthermore, we show that our method is competitive in terms of computational scaling when compared to existing kernel-based ergodic coverage methods~\cite{sun2024fast} while not requiring domain specific information.
    % By employing the ergodic maximum mean discrepancy (EMMD) approach in several problem settings, we show that our method produces highly competitive coverage and information gathering while still being able to retain traditional coverage approaches in complex spaces. 
    Our generalization extends the versatility of ergodic trajectory optimization, enabling future robotic exploration tasks about general objects and environments. 

    In summary, the contribution of this work is a novel formulation of ergodic trajectory optimization that:
    \begin{itemize}
        \item Bases exploration on maximum mean discrepancy.
        \item Extends the versatility of traditional ergodic search to arbitrary domains.
        \item Enables ergodic trajectory optimization on Lie groups and with differential kinematic constraints.
    \end{itemize}

    The paper is outlined as follows: Section~\ref{sec:related_work} overviews related work, Section~\ref{sec:prelims} goes over preliminary background information, Section~\ref{sec:emmd} derives the proposed ergodic trajectory optimization using maximum mean discrepancy, and Sections~\ref{sec:results} and~\ref{sec:conclusion} are results and conclusions.

\section{Related Work} \label{sec:related_work}

    \noindent
    \textbf{Coverage-based Methods.} Coverage planning methods aim to optimize a robot's trajectory to maximize the information gathered within a search space. 
    Conventional coverage methods segment the sample space into grids, where paths that visit each area are calculated using variations of traveling salesperson (TSP) solvers~\cite{laporte1987, applegate2006}.     
    Recent advancements in ergodicity-based methods have demonstrated the ability to produce effective coverage trajectories over continuous domains through the use of optimizing a spectral ergodic metric based on Fourier basis functions~\cite{mathew2011, scott2009capturing, miller2015ergodic}. 
    These methods ensure that a robot will eventually cover a domain so long as the trajectory time approaches infinity~\cite{abraham2020, dong2023} and have demonstrated the ability to create effective coverage paths for a variety of information gathering tasks~\cite{miller2013, dong2023, mavrommati2018, sartoretti2021spectral, lee2024}. 
    More recent work derived kernel-based ergodic metric through the $L^2$-norm, circumventing the computational cost of traditional spectral ergodic metrics~\cite{sun2024fast}. 
    However, while this approach remedied the computational scaling, it requires access to the underlying utility measure defined over the search domain.

    \vspace{0.5em}
    \noindent
    \textbf{Numerically-based Coverage on 3D Surfaces.}
    In many coverage problems, the search domain may not be well-defined and requires mesh-based numerical approximation, motivating the development of alternative numerical methods.
    Previous research has shown that ergodic trajectories can be generated over arbitrary meshes~\cite{ivic2022, bilaloglu2024tactile} through computation of a diffusion-based finite-element approximation to an ergodic velocity field.
    These approaches can be computationally expensive, as they involve using the finite element method to solve a partial differential equation over mesh points at each time step.
    Other methods achieve ergodic trajectories on meshes by computing an ergodic metric through derivation of basis functions via the Laplace-Beltrami operator~\cite{bilaloglu2024tactile}. 
    However, these approaches require computing eigenfunctions as spatial basis functions, which is computationally challenging ~\cite{sharp2020, belkin2008}. 
    In the following sections, we propose a novel formulation of an ergodic metric derived from maximum mean discrepancy~\cite{muandet2017, gretton2012}, eliminating these limitations.

\section{Preliminaries}\label{sec:prelims}

% In this section, we present preliminary information regarding the problem formulation, ergodicity, and ergodic coverage. We also discuss two-sample statistical tests, maximum mean discrepancy, and kernel-mean embedding.

\subsection{Ergodicity and Ergodic Trajectory Optimization}

    Let us define the robot's state in discrete time as $x_t\in \mathcal{X} \subseteq \mathbb{R}^n$.
    Let $\mathbf{x} = \{x_0, \ldots, x_{T-1} \}$ be a trajectory of time-horizon $T \in \mathbb{N}$ by following some control input $u_t \in \mathcal{U} \subseteq \mathbb{R}^m$ for $t = 0, \ldots, T-1$ and dynamics $x_{t+1} = f(x_t, u_t)$ from some initial condition $x_0$, and $f : \mathcal{X} \times \mathcal{U} \to \mathcal{X}$.
    Next, let us define $\omega_t \in \Omega$ as a point in an arbitrary domain $\Omega$ equipped with a function $g : \mathcal{X} \to \Omega$ that projects states $x_t \to \omega_t$, e.g., the forward kinematics where $\Omega=\text{SE}(3)$. 
    \begin{definition} \textit{(Time-Averaged Trajectory Distribution)}
        The time-averaged trajectory distribution of a trajectory $\bx$, 
        is given as 
        \begin{equation}
            \rho_{\bx,T}(\omega) = \frac{1}{T} \sum_{t=0}^{T-1} \delta[\omega- g(x_t)],
            \label{eq:time_averaged_distribution}
        \end{equation}   
        where $\delta$ is a dirac-delta function.
    \end{definition}
    \begin{definition} \textit{(Ergodicity)} \label{def:ergodicity}
        ~\cite[Theorem 6.14]{walters_introduction_2000} A trajectory $\bx = \{ x_t\}_{t=0}^{T-1}$ is said to be ergodic if its time-averaged statistics $\rho_{\bx,T}(\omega)$ at the limit $T\to \infty$ \emph{converges weakly} to the spatial probability measure $\mu : \Omega \to \mathbb{R}^+$. More concretely, 
        \begin{equation}
            \lim_{T\to \infty} \int_\Omega \phi(\omega) d \rho_{\bx, T}(\omega) = \int_\Omega \phi(\omega) d\mu(\omega)
        \end{equation}
        for all continuous functions $\phi \in \mathcal{C}(\Omega)$, where by definition,
        \begin{equation}
            \int_\Omega \phi(\omega) d \rho_{\bx, T}(\omega) = \frac{1}{T}\sum_{t=0}^{T-1} \phi(g \circ x_t).
        \end{equation}
        % is a result of the delta function on trajectories as defined in prior literature~\cite{++}.
    \end{definition}

    We quantify ergodicity in a computable manner by considering ergodic metrics of the form 
    \begin{equation}
        \mathcal{E}_\mu(\mathbf{x}) = \Big\Vert \mathcal{F}[\rho_{\bx,T}] - \mathcal{F}[\mu] \Big\Vert_\mathcal{F}^2
    \end{equation}
    where $\mathcal{F}$ is some functional transform of $\rho$ and $\mu$, and $\Vert \cdot\Vert_\mathcal{F}$ denotes a norm in the transformed space. 

    % Note that as $T\to\infty$, the visitation statistics of the robot in the search domain converge onto the utility measure $\mu$ defined over the space $\mathcal{W}$.  
    % In order to achieve ergodicity, it is common to impose a metric on ergodicity that, when minimized, satisfies ergodicity for $T<\infty$~\cite{++}.
    % In practise, one cannot readily impose a metric on ergodicity (as a result of the delta function) which requires some form of transform leading to the general metric form 
    % \begin{equation}
    %     \mathcal{E}_\mu(\mathbf{x}) = \Big\Vert \mathcal{F}[\rho] - \mathcal{F}[\mu] \Big\Vert_\mathcal{F}
    % \end{equation}
    % where $\mathcal{F}$ is some functional transform of $\rho$ and $\mu$, and $\Vert \Vert_\mathcal{F}$ denotes a norm in the transformed space. 
    In prior work~\cite{mathew2011}, $\mathcal{F}$ is commonly defined by the Fourier transform with the norm specified as a Sobolev norm. 
    More recently, work in~\cite{sun2024fast} derived a kernel-based approximation of the $L^2$-norm ergodic metric using the inner-product of delta functions from the trajectory statistics.
    Each way to measure ergodicity fundamentally has the same intuition: $\mathcal{E}_\mu \xrightarrow{T\to \infty} 0$ if and only if $\mathbf{x}$ is ergodic with respect to $\mu$.

    To achieve ergodic trajectories, one commonly formulates an ergodic trajectory optimization problem. 
    More specifically, given a twice-differentiable ergodic metric $\mathcal{E}_\mu$, one can pose the following optimization
    \begin{align}
        \min_{\substack{\mathbf{x}, \mathbf{u}}}\,\, & \cE_\mu(\mathbf{x})  + \sum_{t=0}^{T-1} \ell(x_t, u_t)\\         \text{subject to } & h_1(x_t, u_t) = 0 \,\, \forall t \in [0, T-1] \nonumber \\ & h_2(x_t, u_t) \le 0 \,\, \forall t \in [0,T-1] \nonumber
    \end{align}
    where $\mathbf{u} = \{u_0, \ldots, u_{T-1} \}$ are control inputs, $\ell : \mathcal{X} \times \mathcal{U} \to \mathbb{R}$ is the running cost, and $h_1, h_2$ are equality and inequality constraints which include dynamics and differential kinematic constraints $x_{t+1} = f(x_t, u_t)$, and any initial and final conditions on states $x_0, x_{T-1}$. 

\subsection{Maximum Mean Discrepancy}

    Maximum Mean Discrepancy (MMD) is a statistical test used to determine the difference between two probability distributions based on samples drawn from the respective distributions. 
    This approach involves embedding each distribution into a Reproducing Kernel Hilbert Space (RKHS) and then computing the discrepancy between their kernel-mean embeddings \cite{gretton2012}. 
    By operating in a (possibly infinite-dimensional) RKHS, MMD is able to capture all relevant statistical properties of the underlying distributions, and as a result has seen strong uptake in statistical and machine learning literature \cite{li2017mmd}.
    
    To define the MMD metric, let us specify $k$ to be a positive-definite kernel on $\Omega$ and $\cH$ be its corresponding RKHS. A kernel is \emph{characteristic} if the \emph{kernel mean embedding} for probability measures $p$ on a compact domain $\Omega$,
    \begin{align}
    p \mapsto \mu_p = \E_{\omega \sim p}[ k(\omega, \cdot)],
    \end{align}
    is injective. 
    Given two probability distributions $p$ and $q$ on $\Omega$, the MMD measures the difference between the distributions by comparing their kernel mean embeddings in the RKHS,
    \begin{align}\label{eq:mmd_definition}
        \MMD_k^2(p, q) = \|\mu_p - \mu_q\|_{\cH}^2.
    \end{align}
    Note that this defines a metric on the space of measures on $\Omega$ when $k$ is characteristic.
    Expanding the MMD metric yields
    \begin{align} \label{eq:expanded_mmd}
        \MMD_k^2(p, q) =& \mathbb{E}_p[k(x, x')] \\ &- 2\mathbb{E}_{p,q}[k(x, y)] + \mathbb{E}_q[k(y, y')], \nonumber
    \end{align} 
    where $x, x' \sim p$, $y,y' \sim q$.
    % \footnote{Note that in this work, one of the measures $p$ is exactly known and specified by the discrete trajectory and does not require i.i.d samples as it is a parameterization of the time-averaged distribution. }.
    Given finite samples $\mathbf{x} = \{x_i\}_{i=1}^N$ and $\mathbf{y} = \{y_i \}_{j=1}^M$, MMD can be empirically estimated by 
    \begin{align}\label{eq:mmd_empirical}
        \MMD_k^2&(p, q) \approx \overline{\MMD}_k^2(\mathbf{x}, \mathbf{y}) = \frac{1}{N^2} \sum_{i=1}^N \sum_{i'=1}^N k(x_i, x_{i'}) \\
        &- \frac{2}{NM} \sum_{i=1}^N \sum_{j=1}^M  k(x_i, y_i) + \frac{1}{M^2} \sum_{j=1}^M \sum_{j'=1}^M k(y_j, y_{j'})  \nonumber.
    \end{align}
    
    In our setting, we use MMD to compare the trajectory distribution $\rho_{\bx}(\omega)$ parameterized by the finite discrete trajectory $\bx = \{\bx_t\}_{t=0}^{T-1}$ with samples $\{ \omega_i\}_{i=1}^M$ of a domain drawn from the utility measure $\mu$. 
    % We can empirically estimate MMD from finite i.i.d.~samples $\{\by_i\}_{i=1}^n$ drawn from $\phi$, given by
    % \begin{align}\label{eq:mmd_empirical}
    % \widehat{\MMD_k}&^2(\rho_\bx, \phi) = \frac{1}{T^2} \sum_{t=1}^T \sum_{t'=1}^T k(\mathbf{x}_t, \mathbf{x}_{t'}) \\
    % &- \frac{2}{nT} \sum_{t=1}^T \sum_{i=1}^n  k(\mathbf{x}_t, \mathbf{y}_i) + \frac{1}{n^2} \sum_{i=1}^n \sum_{i'=1}^n k(\mathbf{y}_i, \mathbf{y}_{i'})  \nonumber.
    % \end{align}
    % We emphasize that $\rho_\bx$ is defined as a discrete distribution, and an empirical approximation via i.i.d.~samples is only necessary to approximate the target distribution $\phi$.

\section{Erogodic Maximum Mean Discrepancy}\label{sec:emmd}

\subsection{MMD Metrizes Ergodicity via Weak Convergence}

    MMD is an appropriate metric for ergodicity as it quantifies weak convergence as defined in Definition~\ref{def:ergodicity}.
    We say that a metric $d$ on probability measures \emph{metrizes weak convergence} if a sequence of measures $\mu_n$ converges weakly to $\mu$ if and only if $\lim_{n \to \infty} d(\mu_n, \mu) = 0$.
    A key result that connects ergodicity and MMD is the fact that, for certain kernels, the MMD metrizes weak convergence \cite{sriperumbudur2010}. 
    
    \begin{theorem}[Weak Convergence, \cite{simon2023}, Theorem 7]
    \label{theorem:weak_convergence}
        Let $\Omega$ be a compact Hausdorff space. A bounded, measurable kernel metrizes the weak convergence of probability measures if and only if it is continuous and characteristic with respect to probability measures on $\Omega$.
    \end{theorem}
    
    This result implies that, on a compact measurement space $\Omega$, any continuous, bounded, and characteristic kernel $k$ can be used to define an ergodic metric function to compare the trajectory statistics $\rho_\bx$ with a target distribution $\mu$ \cite{arbel2019}. In particular, we can use the empirical approximation of MMD from Eq.~\eqref{eq:expanded_mmd} and expand $\rho_\bx$ to obtain
    \begin{align}
        &\overline{\MMD}^2_k(\rho_\mathbf{x}, \mu) = \frac{1}{T^2} \sum_{t=0}^{T-1} \sum_{t'=0}^{T-1} k(g \circ x_t, g \circ x_{t'}) \\
        &- \frac{2}{TM} \sum_{t=0}^{T-1} \sum_{j=1}^M  k(g \circ x_i, \omega_j) + \frac{1}{M^2} \sum_{j=1}^M \sum_{j'=1}^M k(\omega_j, \omega_{j'})  \nonumber,
    \end{align}
    where $\{\omega_j\}_{j=1}^M$ is a collection of i.i.d.~samples from $\mu$.
    Here, we emphasize that $\rho_\bx$ is a discrete probability distribution by definition, and therefore does not need to be approximated. As the final term depends only on samples from the domain, which does not impact trajectory optimization, we define the \emph{ergodic MMD metric (E-MMD)} on $\Omega$ to be 
    \begin{align}\label{eq:e_mmd}
        \mathcal{E}^k_\mu(\bx) = \frac{1}{T^2} &\sum_{t=0}^{T-1} \sum_{t'=0}^{T-1} k(g \circ x_t, g\circ x_{t'}) \nonumber \\
        &- \frac{2}{TM} \sum_{t=0}^{T-1} \sum_{j=1}^M  k(g \circ x_t, \omega_j) .
    \end{align}
    Note that metric information on the domain $\Omega$ is contained within the kernel function $k$. 
    That is, \emph{we can alter the problem definition to account for more general search domains solely by fixing the kernel}. 
    For instance, the kernel function can be defined on Euclidean space where $\Omega = \mathbb{R}^3$ with an $L^2$-norm, in $\mathcal{C}$-space if $g$ is the identity map and $\mathcal{X}$ is joint space, or on a Lie group where $\Omega = \text{SE}(3)$.

\subsection{Ergodic Trajectory Optimization via MMD}

    We now formulate the ergodic trajectory optimization problem via MMD. 
    Assuming that we can generate samples $\{ \omega_j \}_{j=1}^M$ from $\Omega$ based on $\mu$, and a kernel function $k : \Omega \times \Omega \to \mathbb{R}$ and $g : \mathcal{X} \to \Omega$ is provided, the trajectory optimization formulation is defined as 
    \begin{align} \label{prob:emmd}
        \min_{\substack{\mathbf{x}, \mathbf{u}}}\,\, & \mathcal{E}^k_\mu(\bx)  + \sum_{t=0}^{T-1} \ell(x_t, u_t)\\ 
        \text{subject to } & h_1(x_t, u_t) = 0 \,\, \forall t \in [0, T-1] \nonumber \\ 
        & h_2(x_t, u_t) \le 0 \,\, \forall t \in [0,T-1] \nonumber
    \end{align}
    where $\mathcal{E}^k_\mu(\bx)$ is defined in Eq.~\eqref{eq:e_mmd}.
    In practice, we use the Radial Basis Function (RBF) kernel
    \begin{equation}
        k(x, y) = \exp\left(-\frac{\|x-y\|^2}{2\sigma^2}\right),
        \label{eq:gaussian_kernel}
    \end{equation}
    with bandwidth parameter $\sigma$ and where the norm $\Vert \cdot \Vert$ can be generally specified for the specific problem. 
    Note that this kernel satisfies the condition of Theorem~\ref{theorem:weak_convergence}.
    The problem definition in~\eqref{prob:emmd} can be solved with any off-the-shelf solver as MMD is convex (note that any kinematics constraints may make the problem non-convex). Here, we use a variation of a nonlinear conjugate gradient solver with an augmented Lagrange method to handle the constraints~\cite{sorber2012unconstrained, birgin2014practical}.

\begin{figure}[ht!]
    \centering
    \includegraphics[width=\linewidth]{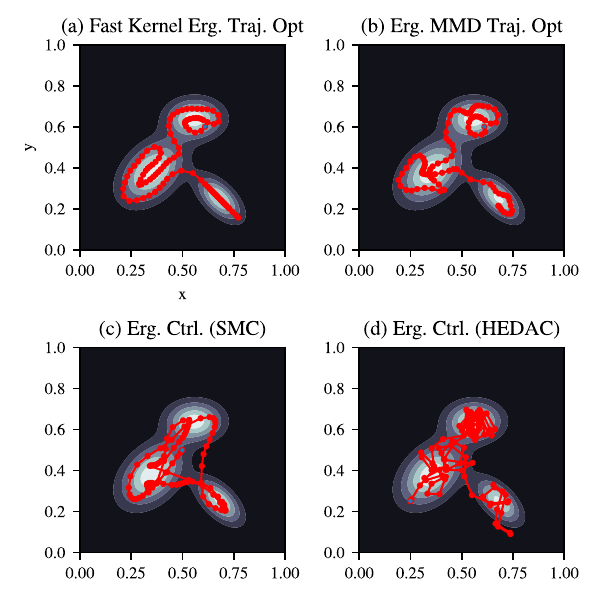}
    \caption{\textbf{Comparison of ergodic search methods.} (a) Ergodic trajectories optimized using the derived kernel via the $L^2$-norm~\cite{sun2024fast}. (b) Ergodic trajectories optimized using MMD (ours) directly on domain samples. Grey areas indicate high-information; both approaches generated comparable levels of ergodic coverage $\mathcal{E}_a=-5.78$, $\mathcal{E}_b-6.23$ (lower is better) according to MMD metric; however our approach did not require gradient information of the underlying distribution. Note discrepancy in ergodicity due to kernel approximation of metric integral over samples. (c-d) Ergodic control methods based on one-step control optimization~\cite{mathew2011,ivic2022}. These approaches require more time to achieve similar levels of ergodicity due to their formulation. (d) Requires computing a partial differential equation at each step on a mesh on the search domain.}
    \label{fig:erg_methods_comparison}
\end{figure}

\section{Results} \label{sec:results}

    \subsection{Trajectory Optimization Pipeline}
        We first describe the procedure of our approach to optimize ergodic trajectories on arbitrary domains, as demonstrated in Fig.~\ref{fig:pipeline}.
        Here, the problem setting is to optimize the trajectory for a robotic manipulator around a cube equipped a utility function that prioritizes coverage of specific cube faces.
        The domain $\Omega$ is specified as the end-effector workspace $\mathcal{W} \subset \text{SE}(3)$ intersected with all points and normals on the surface of the box $\mathcal{B} \subset \text{SE}(3)$, namely $\Omega = \mathcal{W}  \cap \mathcal{B}$, and the state is given as the configuration of the robot $x \in \mathbb{R}^7$ with $u \in \mathbb{R}^7$ as the joint velocity. 
        The function $g : \mathcal{X} \to \text{SE}(3)$ is equipped with a kernel on $\text{SE}(3)$, given by $k(\mathbf{A}, \mathbf{B}) = \exp\left( -\Vert \log \left(\mathbf{A}^{-1} \mathbf{B} \right) \Vert_\Sigma^2 \right)$ for $\mathbf{A,B} \in \text{SE}(3)$ and variance $\Sigma$.
        The kernel measures the distance between two elements in $\text{SE}(3)$ which are identical if $\mathbf{A}^{-1}\mathbf{B}=\mathbf{I}$ where $\mathbf{I}$ is the identity matrix. 
        Using the matrix log projects the $\text{SE}(3)$ product into its Lie algebra $\mathfrak{se}(3)$ which yields a vector space and norm to compute distances~\cite{sun2024fast, chirikjian2012information}. 
        %defined on the Lie algebra $\mathfrak{se}(3)$. 

    \begin{figure}[ht!]
        \centering
        \includegraphics[width=\linewidth]{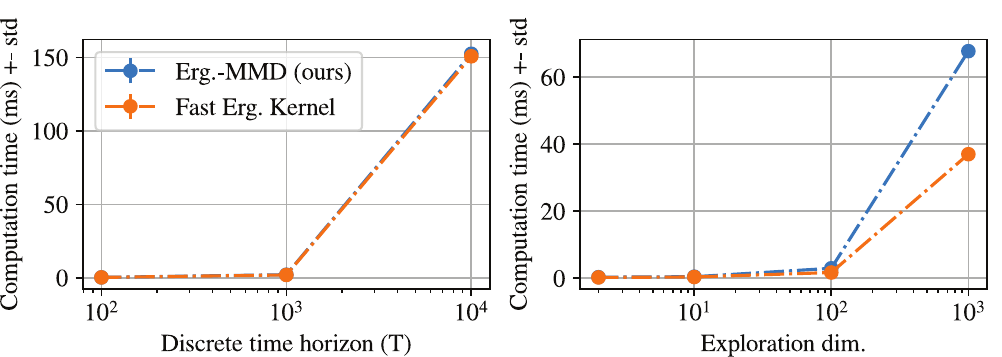}
        \caption{\textbf{Computational scale analysis.} (Left) Computation time with respect to increasing time horizon. (Right) Computation time with respect to increasing state dimension. Comparisons are done with respect to fast ergodic kernel method~\cite{sun2024fast}. Both short comparable performance where our approach is approximately $2\times$ more computationally expensive (due to the kernel approximation of expectations) with respect to state-dimensionality, though the scale order is polynomial for both methods.  }
        \label{fig:computation1}
    \end{figure}
        
        One can practically sample from the domain $\Omega$, e.g., using a depth sensor and a classifier to isolate the sides of interest of the box. 
        In simulation, we sample from $\Omega$ uniformly on the mesh (getting a set of associated points and normals) and use importance sampling to isolate points that have normals pointed to the sides of interest (see Fig.~\ref{fig:pipeline} for illustration).  
        Given the differential constraints of the Franka robot (as joint limits and forward kinematics function) we can solve~\eqref{prob:emmd} directly on the joints and joint velocities which is illustrated in Fig.~\ref{fig:pipeline}. 
        Obstacle avoidance can be naturally integrated by implicitly sampling the domain with an added buffer and integrating the offset in the kernel function.
        
        % The validation of our approach is presented in two stages.
        % First, we demonstrate that our ergodic search method is capable of generating competitive coverage trajectories in simple environments, such as two-dimensional spaces. Second, we show the broader applicability of the proposed EMMD approach by illustrating its ability to produce effective coverage trajectories in complex 3D environments, despite requiring only basic information about the search space and minimal computational resources.

    % \begin{figure}[ht!]
    %     \centering
    %     \includegraphics[width=0.8\linewidth]{figures/emmd_sample_scale.pdf}
    %     \caption{\textbf{Computational scale analysis of sampling.} Here, we show our approach has polynomial computational scaling with respect to the number of samples needed to compute MMD. The Fast Ergodic Search method~\cite{sun2024fast} does not depend on domain samples, thus has constant scaling. However, the Fast Ergodic Search method does not work on arbitrary domains, e.g., mesh-based domains, and requires access to the utility measure $\mu$ and its derivative information, thus showing the trade-off between domain knowledge and having to infer the domain to generate ergodic coverage.}
    %     \label{fig:computation2}
    % \end{figure}

    \subsection{Comparison to Existing Methods}

        Next, we compare our proposed approach in a 2D coverage problem to existing methods for ergodic coverage. 
        We directly compare our work with the Fast Ergodic Search method~\cite{sun2024fast} in the example provided as it is the most similar to our approach.
        The length scale for the kernel was optimized using the method provided in~\cite{sun2024fast} by matching the moments of the kernel means that best captures the underlying utility measure. 
        As additional comparisons, we included the one-step ergodic controllers using SMC~\cite{mathew2011} and HEDAC~\cite{ivic2022}. 
        As shown in Fig.~\ref{fig:erg_methods_comparison}, our proposed approach shows comparable ergodic trajectory to all methods.
        Since each method is using a different ergodic metric, we only quantitatively compare with the Fast Ergodic Search method which has a similar metric that is directly evaluated on the utility measure $\mu$. 
        % A significant advantage of our approach is that the ergodic MMD metric does not require access to underlying utility measure, only samples of the domain. 
        % In contrast, the Fast Ergodic Search method requires direct access to the utility measure and its derivative which is often not available in more complex settings (e.g., meshes). 
        Compared with SMC and HEDAC, the main advantage of our approach is that we solve the ergodic trajectory in a single optimization. 
        SMC and HEDAC compute a one-step control input which considers greedy advancements in ergodicity. 
        As a result, the trajectories require significantly more time to be as ergodic ($\approx 8\times$).
        % TODO ADD IN TIME (HEDAC requires  amount more).
        
        In terms of computation time, we compare our method to Fast Ergodic Search in Fig.~\ref{fig:computation1}. 
        We find that computing the E-MMD metric has comparable computational scaling with the Fast Ergodic Search $L^2$ metric with respect to the time horizon $T$ and the exploration dimension $v$ as shown in Fig.~\ref{fig:computation1} (ours is $\approx 2 \times$ slower in terms of exploration scaling due to the MMD metric approximating the utility measure over samples).
        In particular, our approach for computing the E-MMD metric scales quadratically $\mathcal{O}(M^2)$ with respect to the number of domain samples (where more samples result in a more accurate estimate of the MMD metric).
        % The more samples, the better the estimate of the empirical MMD metric. 
        In contrast,~\cite{sun2024fast} uses domain-specific knowledge (e.g.~the utility measure and its derivative) to avoid the use of sampling, thus eliminating this parameter. 
        % In contrast,~\cite{sun2024fast} does not require samples, thus has constant (or no) scaling with respect to samples, which is achieved with domain specific knowledge, e.g., the utility measure and its derivative. 
        % Note that our approach does not require access to the utility measure and its derivative information where as~\cite{sun2024fast} explicitly requires this information. 
        As a result of requiring this information, the Fast Ergodic Search method cannot be applied on arbitrary domains.
        % , e.g., mesh-based domains, and requires access to the utility measure $\mu$ and its derivative information. 
        Thus, there is a trade-off between the generality of the domain and having to infer the domain to generate ergodic coverage via samples.

\begin{figure}[t!]
    \centering
    \includegraphics[width=\linewidth]{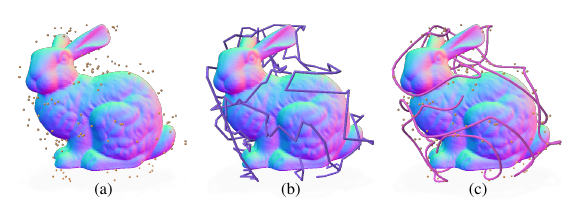}
    \caption{ \textbf{Uniform coverage over bunny.} (a) Uniformly sampled points over bunny mesh. (b) Uniform coverage Traveling Salesperson solution over the sampled points. (c) Uniform ergodic trajectories optimized using MMD (ours) directly on mesh samples. Note that our approach is able to generate uniform coverage trajectories that respect path smoothness constraints globally. }
    \label{fig:bunny_tsp}
\end{figure}

\subsection{Ergodic Search on General Domains}

    Here, we demonstrate a suite of examples of generating ergodic coverage trajectories over various objects.

    \vspace{0.5em}
    \noindent
    \textbf{Search on Bunnies.}
    First, we show that our approach generates smooth trajectories over a bunny domain (see Fig.~\ref{fig:bunny_tsp}). 
    Sample points are generated uniformly about the domain $\Omega$ which is offset for visualization of the trajectories. 
    We compare against solving a traveling salesperson (TSP) path that visits each point sampled at most once. 
    The TSP approach generates paths that traverse the bunny uniformly through a sequential iterative approach~\cite{applegate2006}. 
    Our approach globally solves the trajectory (gradients are with respect to the whole trajectory as opposed to individual points). 
    As a result, we can naturally integrate path constraints, e.g., $\Vert x_{t+1} - x_t \Vert^2$ while still uniformly covering the bunny.

    \begin{figure}[h!]
        \centering
        \includegraphics[width=\linewidth]{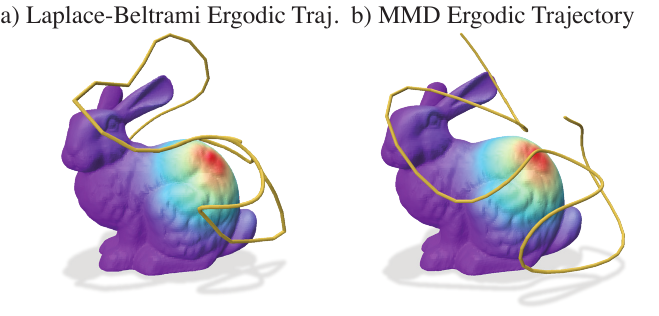}
        \caption{\textbf{Non-uniform coverage over bunny.} (a) Ergodic trajectories optimized using Laplace-Beltrami eigenfunctions ~\cite{bilaloglu2024tactile}. (b) Ergodic trajectories optimized using MMD (ours) directly on mesh vertices. Red areas indicate high-information; both approaches produce comparable coverage where the only difference was the added overhead of computing the basis functions in (a). }
        \label{fig:bunny_laplace}
    \end{figure}

    We additionally compare our approach with a mesh-based approximation of the spectral basis functions using the Laplace-Beltrami operator~\cite{bilaloglu2024tactile} (see Fig~\ref{fig:bunny_laplace}). 
    Here, we include a utility-map on the bunny (red areas indicate higher utility). 
    Both approaches generate qualitatively comparable ergodic trajectories (each approach uses a different definition of ergodicity which makes it challenging to directly compare). 
    The advantage our approach is that 1) we do not need to compute basis functions, and 2) only samples of the search domain are required. 
    In contrast, ~\cite{bilaloglu2024tactile} requires a mesh along with a graph Laplacian to compute an approximation of basis functions to compute an ergodic metric.

    \begin{figure}[ht!]
        \centering
        \includegraphics[trim={0 10cm 0 5cm},clip,width=\linewidth]{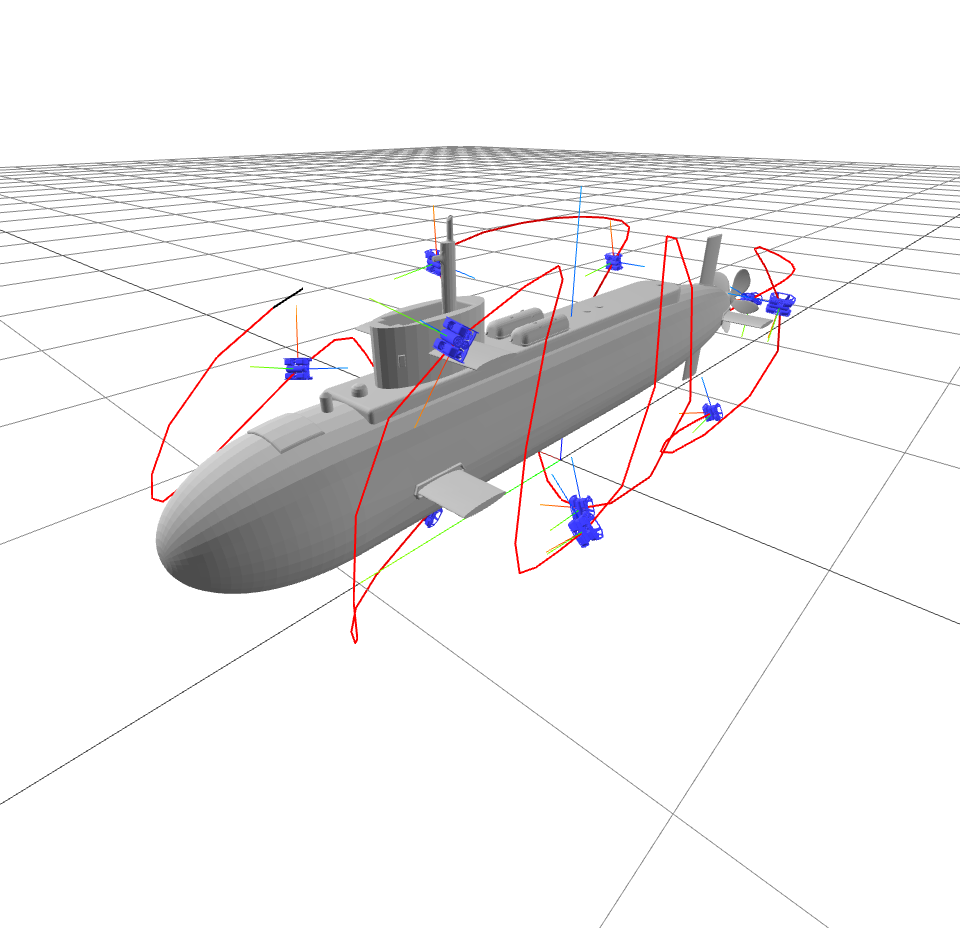}
        \caption{\textbf{Ergodic coverage of submarine over $\mathfrak{se}(3)$. }Ergodic trajectories optimized over $\mathfrak{se}(3)$ using maximum mean-discrepancy metric specify the pose of an underwater vehicle to uniformly inspect the submarine.  }
        \label{fig:submarine}
    \end{figure}
    
\begin{figure}[h]
    \centerline{\includegraphics[scale = 0.27]{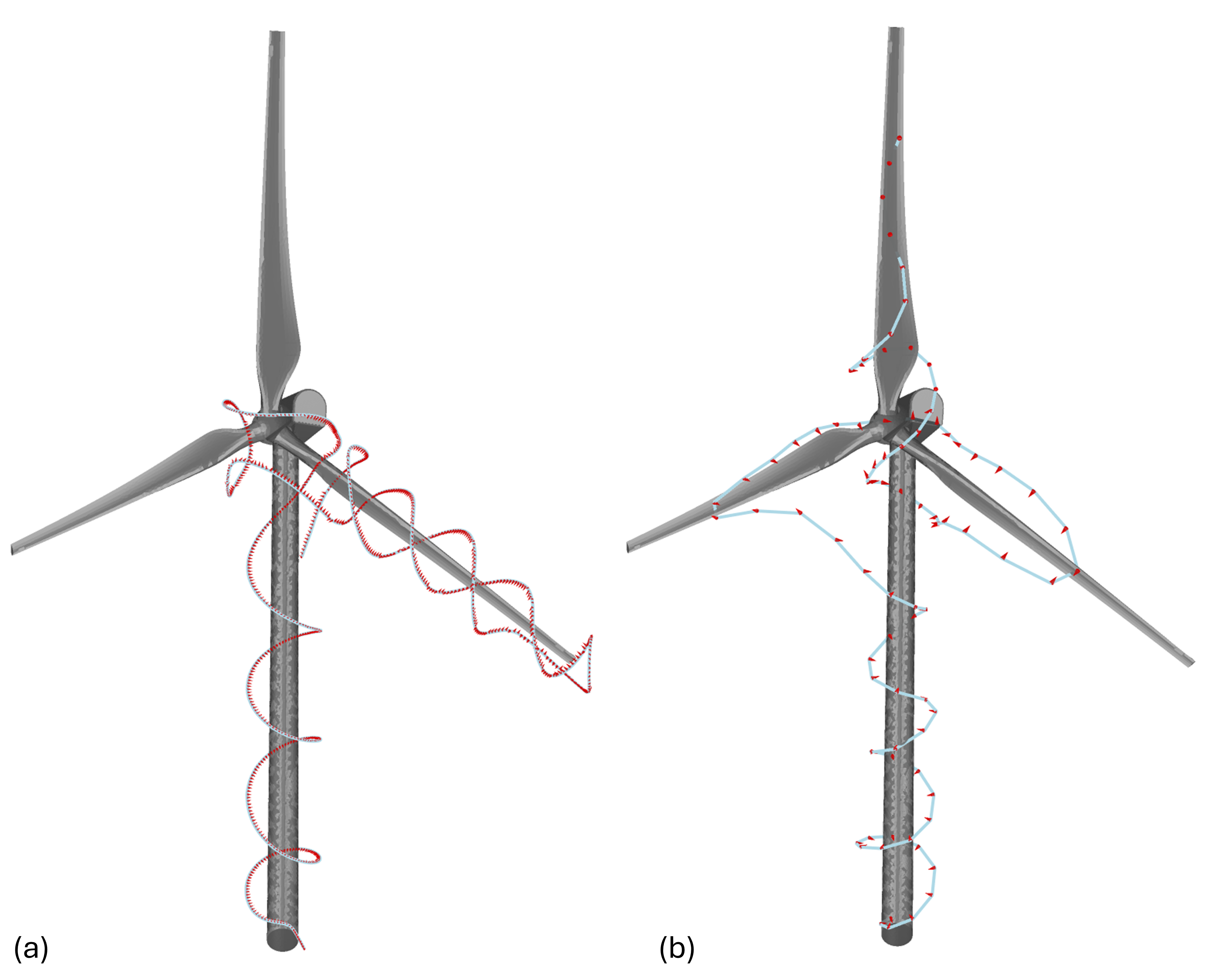}}
    \caption{ \textbf{Ergodic coverage trajectories on a turbine over $\text{se}(3)$.} (a) Ergodic coverage trajectories generated about a uniform distribution using HEDAC~\cite{ivic2022}. (b) Ergodic coverage trajectories generated using our approach via MMD. Since we define the ergodic metric on samples over the domain using MMD, we can globally optimize trajectories to cover the domain. HEDAC requires computing a partial differential equation at each step, making it prohibitively more expensive for planning trajectories as one would need to recompute the PDE continuously.}
    \label{fig:wind_turbine}
    \end{figure}

    \vspace{0.5em}
    \noindent
    \textbf{Coverage on $\mathfrak{se}(3)$.}
    Next, we demonstrate how our approach can be used to directly optimize over Lie groups (specifically, the Lie algebra) to produce coverage paths over arbitrary objects.
    First, we define the search domain to be $\Omega \subset \text{SE}(3)$ which considers only the points and normal orientations on the object domains (here we test a submarine and a wind turbine shown in Fig.~\ref{fig:submarine},~\ref{fig:wind_turbine}). 
    We optimize over the drone's $\mathfrak{se}(3)$ trajectory, that is, $x_t = \xi_t \in \mathfrak{se}(3)$ and the kernel function is defined as $k(\xi_1, \xi_2) = \exp\left( \Vert \log(\exp(\xi_1)^{-1} \exp(\xi_2) \Vert_\Sigma^2 \right)$. 
    With this formulation, we can directly optimize over trajectories in the tangent Lie algebra space $\mathbf{x} = \{ \xi_t\}_{t=0}^{T-1}$ and integrate path constraints of the robot.

    \begin{table}[h]
    \vspace{2.5mm}
    \caption{HEDAC vs. EMMD - Comparison Over Different Cases}
    \label{tab:turbine_comp}
    \begin{tabularx}{\columnwidth}{@{} c c c c c c c c @{}}
    \toprule
    Method & Case & Traj. Length (m) & Covg (\%) & Comp. Time (s) \\ 
    \midrule
    \multirow{3}{*}{HEDAC} & Turbine & 612.00 & 34.211 & 11699.00  \\ 
                           & Portal  & 254.93 & 28.230 & 5229.98  \\ 
                           & Bridge  & 239.64 & 43.548 & 8290.96  \\ 
    \addlinespace
    \multirow{3}{*}{EMMD}  & Turbine & 607.89 & 38.446 & 76.67  \\ 
                           & Portal  & 253.98 & 30.337 & 17.10  \\ 
                           & Bridge  & 203.64 & 72.584 & 236.80 \\ 
    \bottomrule
    \end{tabularx}
    \end{table}

    As demonstrated in Fig.~\ref{fig:submarine} the kernel norm promotes that the robot points towards the surface of the submarine (through an additional transform) while generating uniform coverage throughout the surface of the submarine. 
    We show a similar example for generating coverage on a wind turbine in Fig.~\ref{fig:wind_turbine}. 
    We compare with HEDAC~\cite{ivic2022} which generates ergodic velocity fields through solving a diffusion partial differential equation (PDE) via finite-element method. 
    % We use the $66,360$ points that define the wind turbine mesh vertices to generate the ergodic trajectory. 
    HEDAC uses a one-step control approach to generate the next robot position once the PDE is solved. 
    % once it computes a velocity field based on diffusion at each time-step. 
    As a result, the approach is very myopic, leading to dense coverage at first and only achieving ergodicity as $T\to\infty$. 
    Our approach is able to solve ergodic trajectories over the whole domain at once, spreading coverage throughout the domain. 
    Therefore, to achieve the same levels of ergodicity (calculated using percent coverage in the HEDAC method~\cite{ivic2022}), one would need to run HEDAC for significantly longer time (see Table~\ref{tab:turbine_comp}). 
    Our approach, without any specific optimization on the solver and implementation side, can generate a $T=100$ length trajectory within a minute over $\mathfrak{se}(3)$. 
    The trajectory spreads coverage throughout the wind turbine, where the same $T=100$ length trajectory of HEDAC only covers the stem of the wind turbine.
    Note that our method only covers about $25 \%$ of the wind-turbine as a result of the larger surface area of the turbine. 
    To achieve larger levels of coverage, one would need to run a longer time horizon or add in additional agents to survey the remaining areas.

\section{Conclusion} \label{sec:conclusion}

    In this paper, we introduce a novel approach to ergodic trajectory optimization based on maximum mean discrepancy that extends ergodic trajectory generation to arbitrary domains. 
    We show that MMD is an alternative ergodic metric that can produce effective ergodic trajectories using solely domain samples. 
    Our results demonstrate a wide range of applicability to competitively generate ergodic coverage trajectories across several problem settings with minimal overhead computation. 
    Future work will explore a real-time extension of our approach with time-varying domains.  
    
% Future research will focus on several areas of improvement. While the current EMMD formulation provides effective pre-computed search trajectories, further optimization is required to enable real-time operation in complex environments. We also intend to enhance EMMD’s adaptability by enabling agents to dynamically update their utility distributions as they acquire new information during the search process. Additionally, we intend to optimize the search algorithm for scenarios where the agent’s knowledge of the environment is incomplete at initialization, allowing for incremental updates as the agent explores or gathers more accurate data about its surroundings.


%%%%%%%%%%%%%%%%%%%%%%%%%%%%%%%%%%%%%%%%%%%%%%%%%%%%%%%%%%%%%%%%%%%%%%%%%%%%%%%%

%%%%%%%%%%%%%%%%%%%%%%%%%%%%%%%%%%%%%%%%%%%%%%%%%%%%%%%%%%%%%%%%%%%%%%%%%%%%%%%%

%%%%%%%%%%%%%%%%%%%%%%%%%%%%%%%%%%%%%%%%%%%%%%%%%%%%%%%%%%%%%%%%%%%%%%%%%%%%%%%%

\section*{Acknowledgements}

HW and FR are supported by The University of Sydney. FR is supported by NVIDIA. DL was supported by the Hong Kong Innovation and Technology Commission (InnoHK Project CIMDA). IA and CH are supported by Yale University.

%%%%%%%%%%%%%%%%%%%%%%%%%%%%%%%%%%%%%%%%%%%%%%%%%%%%%%%%%%%%%%%%%%%%%%%%%%%%%%%%

\bibliographystyle{IEEEtran}
\bibliography{references}

\end{document}